\documentclass[10pt,twocolumn,letterpaper]{article}

\usepackage{cvpr}              

\usepackage[dvipsnames]{xcolor}

\usepackage{multirow}
\usepackage{colortbl}
\usepackage{pifont}
\usepackage{adjustbox}

\definecolor{cvprblue}{rgb}{0.21,0.49,0.74}
\usepackage[pagebackref,breaklinks,colorlinks,citecolor=cvprblue]{hyperref}

\def\paperID{17354} 
\def\confName{CVPR}
\def\confYear{2024}

\title{DiffAugment: Diffusion based Long-Tailed Visual Relationship Recognition}
\author{Parul Gupta\\
Monash University\\
{\tt\small parul@monash.edu}
\and
Tuan Nguyen\\
Monash University\\
{\tt\small tuan.ng@monash.edu}
\and
Abhinav Dhall\\
Indian Institute of Technology Ropar\\
{\tt\small abhinav@iitrpr.ac.in}
\and
Munawar Hayat\\
Monash University\\
{\tt\small munawar.hayat@monash.edu}
\and
Trung Le\\
Monash University\\
{\tt\small trunglm@monash.edu}
\and
Thanh-Toan Do\\
Monash University\\
{\tt\small toan.do@monash.edu}
}

\begin{document}
\maketitle
\begin{abstract}
The task of Visual Relationship Recognition (VRR) aims to identify relationships between two interacting objects in an image and is particularly challenging due to the widely-spread and highly imbalanced distribution of $<$subject, relation, object$>$ triplets. To overcome the resultant performance bias in existing VRR approaches, we introduce DiffAugment -- a method which first augments the tail classes in the linguistic space by making use of WordNet and then utilizes the generative prowess of Diffusion Models to expand the visual space for minority classes. We propose a novel hardness-aware component in diffusion which is based upon the hardness of each $<$S,R,O$>$ triplet and demonstrate the effectiveness of hardness-aware diffusion in generating visual embeddings for the tail classes. We also propose a novel subject and object based seeding strategy for diffusion sampling which improves the discriminative capability of the generated visual embeddings. Extensive experimentation on the GQA-LT dataset shows favorable gains in the subject/object and relation average per-class accuracy using Diffusion augmented samples.Our code will be released.
\end{abstract}
\section{Introduction}
\label{sec:intro}

The Long-tailed Visual Relationship Recognition (LTVRR) task aims at understanding the relationships among interacting objects in an image when both the objects and the relationships follow a long-tailed distribution. Since Diffusion Models (DM) \cite{ho2020denoising} have proven to be quite effective in modelling large-scale complex visual-textual space which is naturally long-tailed \cite{liu2019largescale}, this work aims at utilizing the generative capabilities of Diffusion Models to overcome the performance bias arising due to long-tail specifically for Visual Relationship Recognition.

The goal of the VRR task is to recognize the categories of two interacting objects and their relation, \eg recognizing triplets like $<$banans, hang from, ceiling$>$. Due to the enriched scene understanding provided by VRR, it benefits various other vision tasks such as image captioning, Visual Question-Answering and 3D scene synthesis. However, due to the imbalanced class distribution in many VRR datasets, predictions of the most existing models are dominated by the head/frequent relations, lacking generalization on tail/low-shot relationships.
In the LTVRR setup which was introduced first in \cite{abdelkarim2021exploring}, subjects, objects and relations follow a long-tailed distribution. They introduced two benchmarks -- dubbed VG8K-LT and GQA-LT built upon the widely used Visual Genome \cite{krishna2016visualgenome} and GQA datasets \cite{hudson2019gqa} respectively. These benchmarks are particularly challenging as not only could the combination (S, R, O) be rare, but so can any of the interacting subjects/objects (S/O) and/or the relation (R). Moreover, they have extremely high imbalance ratios for both objects and relationships -- for GQA-LT, the ratio of the number of samples in the most populated to the least populated object class is 300,000+ and for relation classes, the ratio is around 1.7 million. Similarly for VG8K-LT, the imbalance ratio is 14,000 for object classes and 34,000 for relation classes.

A general VRR approach consists of 3 steps -- (1) Extracting general visual features of cropped image regions of the subject, object and relation using pretrained object detection models such as Faster-RCNN \cite{ren2016faster}. These features can be considered as the standard input to any VRR approach. (2) Refining the visual features by message-passing among objects/relations. (3) Predicting the categories of subject, object and corresponding relation using the refined embeddings.
The standard strategies to tackle long-tailed imbalance are based upon data re-sampling \cite{shen2016relay} or weight adjustment based loss functions such as weighted cross-entropy, focal loss \cite{lin2018focal}, equalization loss \cite{tan2020equalization} or decoupling \cite{kang2020decoupling} the representation learning from classifier learning. Another direction for alleviating imbalance involves augmentation of the tail classes, wherein features learnt from the abundantly populated head classes are transferred to the under represented tail classes \cite{liu2019largescale}. Building upon these approaches, for Visual Relationship Recognition in the long-tail setting, \cite{abdelkarim2021exploring} introduces a Visio-Lingual Hubless loss and RelMix augmentation. Both RelMix and VilHub loss can be integrated with the 
class-balancing loss functions and decoupling to produce good results.


Complementary to these strategies, we propose DiffAugment: an approach which utilizes Diffusion Models \cite{ho2020denoising} to augment the tail classes and overcome the imbalance. Unlike RelMix augmentation which operates on the refined visual embeddings, DiffAugment works at the VRR input level. To do so, firstly, we augment the triplets containing objects/relations from tail classes in the training data. For this, we take the triplets containing tail objects/relations from training data and replace their subjects/objects with similar object classes from the dataset. We make use of Wordnet \cite{10.1145/219717.219748} based LCH synset similarity \cite{10.7551/mitpress/7287.003.0018} to get the similar object classes. Parallely, we also train a Diffusion Model that generates the visual features of step 1 (general VRR input), conditioned upon the CLIP \cite{radford2021learning} textual embedding of the triplets. Next, we sample the visual features corresponding to the augmented triplets from this trained DM, which in turn, can be used to fine-tune any existing LTVRR model. In order to improve the quality of the generated visual features, we further propose two enhancements in the Diffusion Model-- (1) Motivated from \cite{samuel2023seedselect}, rather than starting from a purely random Gaussian distribution \textit{(seed)} while sampling, we extract the visual features of the augmented triplet's subject and object from the training data to obtain a better \textit{seed}. (2) We model the \textit{hardness} of a triplet and use it as an extra condition while training the Diffusion Model. This \textit{Hardness} can be interpreted as a constraint on the region in which the generated visual embedding of a triplet can occur, thereby improving the discriminative capability of the generated visual features. Finetuning using our generated visual features shows consistent improvement in the per-class accuracy for different LTVRR approaches on the GQA-LT dataset.\\
Thus our contributions in this work can be summarized as:
\begin{itemize}[noitemsep, nolistsep]
\item To the best of our knowledge, this work is the first attempt to employ Diffusion Models in the domain of Visual Relationship Recognition. 
\item Our approach successfully models and generates the visual embeddings of captions (triplets) involving tail classes and uses them to overcome the bias in the training of existing Visual Relationship Recognition approaches. Hence, it acts as a data augmentation strategy which can be used on top of any existing Visual Relationship Recognition algorithm to improve the classification performance on the tail classes.
\item  We introduce two novel components - a Subject/Object based seeding strategy and \textit{hardness-aware} Diffusion wherein we define the \textit{hardness} of each triplet and add it as a condition to the Diffusion Model - both of these components improve the discriminative capability of the DM generated samples.
\end{itemize}

\section{Related Work}
\label{sec:relwork}
\noindent\textbf{Long-tailed Visual Relationship Recognition}
There are several works pertaining to Visual Relationship Recognition in long-tailed data distribution. The first work in this direction, \cite{abdelkarim2021exploring} introduced a novel augmentation strategy called RelMix and a visio-linguistic hubless (VilHub) loss to adapt a base approach for Visual Relationship Recognition (LSVRU \cite{zhang2019largescale}) to the long-tail scenario. RelMix is inspired from Manifold Mixup \cite{verma2019manifold} and combines the visual embeddings of different triplets belonging to tail/medium classes along with their expected predictions to augment the training data. VilHub loss encourages the average probability of each class being predicted in a batch to be close to uniform, \ie equally preferred across head and tail. While our approach is essentially an augmentation strategy too, it employs a Diffusion Model to generate the foundation Faster-RCNN \cite{ren2016faster} based visual features rather than the approach-specific refined features and can be used on top of any model trained using RelMix and VilHub. RelMix is an end-to-end augmentation strategy and needs a VRR model to be trained from scratch whereas our diffusion based augmentation needs only fine-tuning of already trained VRR models.\\
Another state-of-the-art work for LTVRR titled RelTransformer \cite{chen2022reltransformer} uses two Transformer \cite{vaswani2023attention} based encoders - one to represent the global scene context and another for the relation. The global scene context encoder guides the relation encoder through meshed attention. Further, it also uses a novel memory attention module to allow the relation representation information to be shared across the entire dataset to alleviate the frequency bias. ViTSCG \cite{wang2023vtscg} uses Vision Transformer \cite{dosovitskiy2021image} and Masking with overlapped area (MOA) module \cite{park2023viplo} to first extract the subject and object features. Now, to obtain the relation features, it uses spatially conditioned graph \cite{zhang2021spatially} which is a graph neural network designed to jointly reason about the appearance and spatial information of an image. Finally, the subject, object and relation features are refined using the RelTransformer module \cite{chen2022reltransformer} as explained earlier and passed on to the MLP classifiers for prediction. Our DiffAugment strategy can be utilised for both these approaches by augmenting the features that are input to the RelTransformer module.\\
\noindent\textbf{Diffusion Models} (DM) \cite{ho2020denoising} are a recently introduced class of generative models which have shown promising results in numerous generative applications including image super-resolution \cite{saharia2021sr3}, image editing \cite{brooks2022instructpix2pix, meng2022sdedit}, speech synthesis \cite{kong2020diffwave, lee2021priorgrad, huang2022fastdiff}, voice conversion \cite{popov2021diffusion} and text-to-speech \cite{popov2021grad}. The basic DM architecture has been improved in various ways recently with attempts to improve the diversity and quality of the generated images while reducing the training and sampling time \cite{zhang2023fast, zheng2023fast, li2023qdiffusion, tang2023improved}. Some recent works such as Class-Balancing Diffusion Model (CBDM) \cite{qin2023classbalancing} also venture towards maintaining the sample quality and diversity when training Diffusion Models on imbalanced datasets. In this paper, we make use of the simplest version of Diffusion Models called Latent Diffusion Models with Cross-Attention based conditioning \cite{rombach2022highresolution}. Improvements in LTVRR by using advanced Diffusion Models has been left to be explored in the future.\\
\section{Method}
\label{sec:method}
\begin{figure*}
\begin{center}
\includegraphics[scale=0.3]{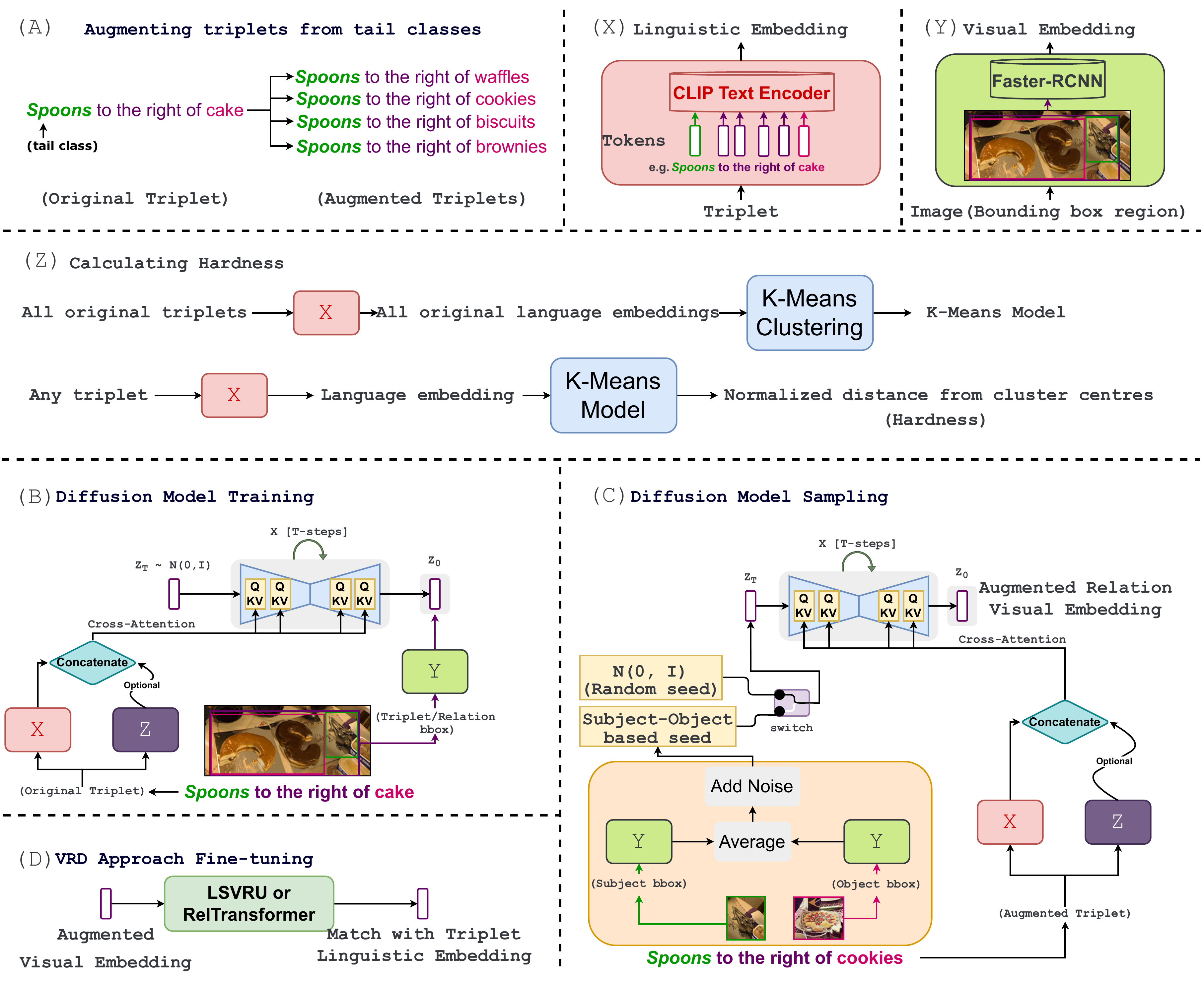}
\end{center}
   \caption{\textbf{For all stages: }X) Given a triplet $<$S,R,O$>$, its linguistic embedding is obtained using the CLIP \cite{radford2021learning} text encoder. Y) For any bounding box region from an image, its visual embedding is obtained from VGG-16 \cite{simonyan2015deep} architecture based Faster-RCNN \cite{ren2016faster} backbone. Z) A K-means cluster model is trained on the CLIP linguistic embeddings of all the original triplets. Using this model, the Hardness based conditional of any triplet is the normalised vector of distances between K-means cluster centres and the triplet's CLIP language embedding. \textbf{DiffAugment Pipeline:} A) In stage 1, we augment the triplets containing s/r/o from tail classes using Wordnet \cite{10.1145/219717.219748} based LCH Synset similarity \cite{10.7551/mitpress/7287.003.0018}. B) In stage 2, the Diffusion Model is trained in the visual embedding space, conditioned upon the linguistic embedding and optionally concatenated with the hardness conditional. C) The trained Diffusion Model is used to sample visual embeddings corresponding to the augmented triplets of stage 1, optionally using a subject and object based starting seed. D) The augmented visual features can be used to fine-tune any pre-trained Visual Relationship Detection model, such as LSVRU \cite{zhang2019largescale} or RelTransformer \cite{chen2022reltransformer}.}
\label{fig:LTVRD_pipeline}
\end{figure*}
\subsection{Problem Definition}
In the Visual Relationship Recognition task, each image $I$ is assumed as a scene graph $G=(N,E)$ where each node $n_i \in N$ represents an object in the image and each edge $e_i \in E$ represents the spatial or semantic relation between two interacting objects. The visual relationship between a subject $s$ and an object $o$ is denoted by $r$. The goal is to predict the labels $y_r$, $y_s$ and $y_o$ (corresponding to the relation, subject and object respectively), given the Image $I$ and the bounding boxes $b_s$ and $b_o$ corresponding to the subject and object respectively.
\begin{equation*}
    y_s, y_r, y_o = f(b_s, b_o, b_r, I)
\end{equation*}
The bounding box corresponding to the relation $b_r$ is obtained by the minimum enclosing region of $b_s$ and $b_o$. $I$ is the raw RGB pixels of the image and $f$ denotes the inference function.

\subsection{Diffusion-based Augmentation -- DiffAugment}
\label{subsec:DiffAugment}
 Our base strategy has four different stages which are described here. Figure \ref{fig:LTVRD_pipeline}(A, B, C, D) show the overall pipeline.
\paragraph{1. Augmenting the Triplets.} (Shown in Figure \ref{fig:LTVRD_pipeline}(A)) For each original triplet involving a tail relation, we replace the subject/object with similar classes from the training data. Likewise, for each original triplet involving a tail subject/object, we replace the corresponding object/subject respectively with similar classes from the training data. In order to obtain the similar classes, we use Wordnet \cite{10.1145/219717.219748} based Leacock Chordorow (LCH) \cite{10.7551/mitpress/7287.003.0018} Synset similarity metric which ensures that the augmented triplets are plausible examples. \eg In the Figure \ref{fig:LTVRD_pipeline}(A), the triplet $<$spoons, to the right of, cake$>$ involves the tail class \textit{spoons} as the subject. Therefore we replace the object, \ie \textit{cake} with other object classes from the dataset that are similar to \textit{cake}, such as \textit{waffles}, \textit{cookies}, \textit{biscuits} and \textit{brownies}.
\paragraph{2. Diffusion Model Training.} (Shown in Figure \ref{fig:LTVRD_pipeline}(B)) We choose the VGG-16 \cite{simonyan2015deep} backbone based Faster-RCNN \cite{ren2016faster} visual features (also used in LSVRU \cite{zhang2019largescale}, RelTransformer \cite{chen2022reltransformer}) as the space for Diffusion (also shown in Figure \ref{fig:LTVRD_pipeline}(Y)). First, we extract the Faster-RCNN features of the bounding boxes corresponding to the relations (\ie $b_r$) for all the triplets in the training data. Then, we use these visual features as the input and target for the forward and backward processes of Diffusion. The conditioning of the Diffusion Model is done upon the CLIP textual encoding of the triplets (denoted by (X) in Figure \ref{fig:LTVRD_pipeline}). To condition the Diffusion Model on the textual embeddings, cross-attention mechanism \cite{rombach2022highresolution} is applied. As an enhancement, we also use a hardness based conditional for the Diffusion Model, which is explained in Section \ref{subsec:hardness_aware_diffusion}.
\paragraph{3. Diffusion Model Sampling.} (Shown in Figure \ref{fig:LTVRD_pipeline}(C)) The trained Diffusion Model is used to sample the Faster-RCNN \cite{ren2016faster} based visual features for the triplets augmented in the first stage. For example, in Figure \ref{fig:LTVRD_pipeline}(C), the CLIP embedding of the augmented triplet $<$spoons, to the right of, cookies$>$ is used as a conditional input to sample the Faster-RCNN visual embedding corresponding to it. Instead of using a random Gaussian seed for sampling, we can also use a seed obtained using the augmented triplet's subject and object, as explained in Section \ref{subsec:s_o_seed}.
\paragraph{4. VRD Approach Fine-tuning.} (Shown in Figure \ref{fig:LTVRD_pipeline}(D)) The generated visual features can be used to fine-tune any existing, pre-trained Visual Relationship detection model since most of the existing VRR approaches use Faster-RCNN \cite{ren2016faster} based features as an input. We perform fine-tuning on two such VRD methods - LSVRU and RelTransformer and observe an improvement in the average per class accuracy for both.
\subsection{Enhancement-1: Subject-Object based seed}
\label{subsec:s_o_seed}
Many image-editing based applications of Diffusion Models make use of the noisy versions of original images when trying to generate the modified image \cite{Avrahami_2023}. Inspired from this idea, while generating the visual features of any augmented triplet, we propose to make use of the visual features of its subject and object. To do so, for each augmented triplet, we choose random bounding boxes corresponding to its subject and object from the training data and take an average of their Faster-RCNN \cite{ren2016faster} features. Then we add Gaussian noise to it in the same manner as the forward diffusion process and use the result as the starting seed while sampling from the trained Diffusion Model. This is shown in Figure \ref{fig:LTVRD_pipeline}(C), where bounding boxes corresponding to the subject \textit{spoons} and object \textit{cookies} are taken randomly from the training dataset; following which their Faster-RCNN features are combined to produce the Subject-Object based seed. As observed in Section \ref{sec:experiments}, samples generated in this manner have a better discriminative quality.
\subsection{Enhancement-2: Hardness aware Diffusion}
\label{subsec:hardness_aware_diffusion}
As the vocabulary of the data on which the CLIP \cite{radford2021learning} textual encoder is trained is much larger than the limited number of classes in the GQA-LT dataset (1703 objects and 310 relations), the linguistic embeddings used to condition the Diffusion Model are not as discriminative as desirable. Therefore, we propose to generate an additional, hardness based conditional for the Diffusion Model. For this, we take the CLIP \cite{radford2021learning} linguistic embeddings of all the triplets in the training dataset and perform K-means clustering. To obtain the hardness of any triplet, we calculate the distance of its CLIP textual \cite{radford2021learning} embedding from all the cluster centres of the K-Means model and normalize the resulting vector of distances. This process is displayed in Figure \ref{fig:LTVRD_pipeline}(Z). Just like the memory augmentation vectors in RelTransformer \cite{chen2022reltransformer} encode the global relation information for the whole dataset, similarly, the cluster centres of the CLIP textual embeddings encode information about the entire dataset, \ie they are a compressed representation of the entire dataset. 
Intuitively, classes that are similar to each other are expected to have similar visual features. Therefore, the cluster centres in the CLIP textual space can be assumed to correspond to the cluster centres in the Faster-RCNN visual space. 
So we expect that the generated visual embedding of any triplet should be closer to the visual embeddings of those triplets whose linguistic embeddings are similar to that of the current triplet.
To encourage that, we use the \textit{hardness} vector as a condition in the Diffusion Model, in addition to the CLIP textual embedding.  
Specifically, our \textit{hardness} vector 
encodes the distances of a triplet with respect to the textual cluster centres.
 Hence, at sampling time, (Figure \ref{fig:LTVRD_pipeline}(C)), this vector explicitly encourages the Diffusion Model to generate its visual embedding at similar relative distance from the corresponding visual cluster centres. Ablations in Section \ref{sec:experiments} demonstrate the effectiveness of introducing this hardness based conditional to the Diffusion Model. 
\section{Experiments}
\label{sec:experiments}
\subsection{Experimental settings}
In this work, we perform experiments on the \textbf{GQA-LT} dataset built on top of Visual Genome by \cite{abdelkarim2021exploring}. It has $72,580$ training images, $2,573$ validation images and $7,722$ test images; with $1,703$ objects and $310$ relations. The most frequent object and relation has $374,282$ and $1,692,068$ examples,
and the least frequent has $1$ and $2$ respectively. Following the strategy used by \cite{abdelkarim2021exploring}, the dataset is split into three parts-- Many, Medium, Few and the selection ratio of each split is based on the frequency of each class: Many (top $5\%$ - 86 classes for S/O and 16 classes for R), Medium(middle $15\%$ -  255 classes for S/O and 46 classes for R), Few (remaining $80\%$ - 1362 classes for S/O and 248 classes for R).\\
\noindent\textbf{Baseline Models} We evaluate the effect of our diffusion based augmentation on two state-of-the-art Visual Relationship Recognition approaches -- LSVRU \cite{zhang2019largescale} and RelTransformer \cite{chen2022reltransformer}.
For both the baselines, we also consider models trained using class-balancing loss functions such as weighted cross-entropy loss (WCE) for fine-tuning. For both the methods, since trained model checkpoints are not publicly available, we train the models from scratch using their official implementations\footnote{\url{https://github.com/Vision-CAIR/RelTransformer}, \url{https://github.com/Vision-CAIR/LTVRR/tree/ltvrd-challenge-2023}} in order to get baseline results\footnote{We are unable to reproduce the performance reported for RelTransformer with WCE loss using the official implementation. Hence we report the fine-tuning results based upon the performance that we get in Table \ref{Table:Quantitative_Results}.} and further fine-tune them on our generated samples.\\
\noindent\textbf{Evaluation metrics} The main metric used is the average per-class accuracy, which is the accuracy of each class calculated separately, then averaged.\\
\noindent\textbf{Implementation details} For augmenting the triplets in stage 1 (Section \ref{subsec:DiffAugment}), we consider the triplets containing either few or medium classes and augment with synsets having LCH similarity greater than or equal to 2.26 (empirical value). The CLIP text encoder produces a 768-D linguistic embedding while the Faster-RCNN output is a 4096-D visual embedding. In order to calculate the hardness of each triplet, K-means clustering is performed with 1200 cluster centres, thus giving a 1200-D hardness vector for each triplet after L1 normalization. A total of 48K augmented triplets from few classes and 48K augmented triplets from medium classes are used for the experiments. While the training of the baselines required 8 Nvidia V100 GPUs with a batch size of 8 and is done for 12 epochs, finetuning can be performed on a single GPU. We use a batch size of 256 for the augmented visual embeddings' data loading and fine-tune for 10 epochs on the augmented data.
\begin{table*}
\begin{center}
\begin{adjustbox}{max width=\textwidth}
\begin{tabular}{l|cccc|cccc|c}
\toprule
\multirow{3}{*}{Learning Methods} & \multicolumn{4}{c|}{Subject/Object} & \multicolumn{4}{c|}{Relation} & Combined\\
& many & medium & few & all & many & medium & few & all & all\\
& 86 & 255 & 1362 & 1703 & 16 & 46 & 248 & 310 & 2013\\
\midrule
\multicolumn{10}{c}{Architecture: LSVRU \cite{zhang2019largescale}}\\
\hline
CE & 68.24 & 36.34 & 6.74 & 14.28 & 60.24 & 13.70 & 6.34 & 10.22 & 12.25\\
\cellcolor{lightgray!50}CE + DiffAugment & \cellcolor{lightgray!50}62.23 & \cellcolor{lightgray!50}41.77 & \cellcolor{lightgray!50}10.14 & \cellcolor{lightgray!50}17.51 & \cellcolor{lightgray!50}35.21 & \cellcolor{lightgray!50}23.94 & \cellcolor{lightgray!50}8.04 & \cellcolor{lightgray!50}11.80 & \cellcolor{lightgray!50}14.66\\
CE + VilHub + RelMix & \underline{68.80} & \underline{44.20} & 10.26 & 18.30 & \underline{63.86} & 12.00 & 6.83 & 10.55 & 14.42\\
\cellcolor{lightgray!50}CE + VilHub + RelMix + DiffAugment & \cellcolor{lightgray!50}62.66 & \cellcolor{lightgray!50}44.08 & \cellcolor{lightgray!50}12.79 & \cellcolor{lightgray!50}19.99 & \cellcolor{lightgray!50}48.94 & \cellcolor{lightgray!50}19.20 & \cellcolor{lightgray!50}7.89 & \cellcolor{lightgray!50}11.69 & \cellcolor{lightgray!50}15.84\\
WCE & 54.83 & 43.32 & 12.25 & 19.05 & 52.75 & 35.17 & 13.03 & 18.37 & 18.71\\
\cellcolor{lightgray!50}WCE + DiffAugment & \cellcolor{lightgray!50}39.83 & \cellcolor{lightgray!50}33.90 & \cellcolor{lightgray!50}\underline{17.33} & \cellcolor{lightgray!50}\textbf{20.95} & \cellcolor{lightgray!50}37.37 & \cellcolor{lightgray!50}\underline{37.72} & \cellcolor{lightgray!50}\underline{18.13} & \cellcolor{lightgray!50}\textbf{22.03} & \cellcolor{lightgray!50}\textbf{21.49}\\
\hline
\multicolumn{10}{c}{Architecture: RelTransformer \cite{chen2022reltransformer}}\\
\hline
CE & \underline{72.45} & 50.06 & 11.69 & 20.50 & \underline{62.48} & 16.83 & 7.45 & 11.69 & 16.10\\
\cellcolor{lightgray!50}CE + DiffAugment & \cellcolor{lightgray!50}70.55 & \cellcolor{lightgray!50}\underline{50.47} & \cellcolor{lightgray!50}13.23 & \cellcolor{lightgray!50}21.70 & \cellcolor{lightgray!50}47.89 & \cellcolor{lightgray!50}30.13 & \cellcolor{lightgray!50}7.09 & \cellcolor{lightgray!50}12.62 & \cellcolor{lightgray!50}17.16\\
WCE & 53.67 & 47.35 & 19.47 & 25.37 & 54.86 & 38.96 & 15.10 & 20.70 & 23.04\\
\cellcolor{lightgray!50}WCE + DiffAugment & \cellcolor{lightgray!50}39.16 & \cellcolor{lightgray!50}38.04 & \cellcolor{lightgray!50}\underline{22.68} & \cellcolor{lightgray!50}\textbf{25.81} & \cellcolor{lightgray!50}45.19 & \cellcolor{lightgray!50}\underline{40.19} & \cellcolor{lightgray!50}\underline{18.01} & \cellcolor{lightgray!50}\textbf{22.71} & \cellcolor{lightgray!50}\textbf{24.26}\\
\bottomrule
\end{tabular}
\end{adjustbox}
\end{center}
\caption{\textbf{Average per-class accuracy on GQA-LT dataset}. The baseline results (non-gray) are as per our reproduction using the official code released by the authors \cite{chen2022reltransformer, abdelkarim2021exploring}. The DiffAugment results include both the enhancements, \ie using Subject-Object based seed and Hardness aware diffusion. The overall best results are in bold, while category-wise best results are underlined. Combined accuracy refers to the average of all subjects/objects accuracy and all relations accuracy. Diffusion based augmentation improves the average per class accuracy for both LSVRU and RelTransformer; with or without class-balancing weighted cross entropy loss, VilHub loss and RelMix augmentation.}
\label{Table:Quantitative_Results}
\end{table*}
\begin{table*}
\begin{center}
\begin{tabular}{l|l|cccc|cccc|c}
\toprule
\multirow{3}{*}{Learning Method} & \multirow{3}{*}{Seed} & \multicolumn{4}{c|}{Subject/Object} & \multicolumn{4}{c|}{Relation} & Combined\\
& & many & medium & few & all & many & medium & few & all & all\\
& & 86 & 255 & 1362 & 1703 & 16 & 46 & 248 & 310 & 2013\\
\midrule
\rowcolor{lightgray!50}\multicolumn{11}{l}{LSVRU \cite{zhang2019largescale}}\\
\multirow{2}{*}{CE + DiffAugment} &  Random & 62.12 & \underline{42.42} & \underline{10.18} & \textbf{17.63} & 34.67 & 23.73 & 6.04 & 10.14 & 13.88\\
& S-O & \underline{62.23} & 41.77 & 10.14 & 17.51 & \underline{35.21} & \underline{23.94} & \underline{8.04} & \textbf{11.80} & \textbf{14.66}\\
\hline
\rowcolor{lightgray!50}\multicolumn{11}{l}{RelTransformer \cite{chen2022reltransformer}}\\
\multirow{2}{*}{WCE + DiffAugment} &  Random & \underline{39.49} & 38.02 & \underline{22.92} & \textbf{26.02} & \underline{46.79} & 39.59 & 17.27 & 22.10 & 24.06\\
& S-O & 39.16 & \underline{38.04} & 22.68 & 25.81 & 45.19 & \underline{40.19} & \underline{18.01} & \textbf{22.71} & \textbf{24.26}\\
\bottomrule
\end{tabular}
\end{center}
\caption{\textbf{Effect of using Subject-Object based (S-O) seed} rather than Random Gaussian seed for diffusion sampling. Combined accuracy refers to the average of all subjects/objects accuracy and all relations accuracy. For all the experiments, Hardness aware diffusion has been used. The overall better results are in bold. Category-wise better results are underlined. Sampling using subject-object based seed gives better combined performance.}
\label{Table:Ablation_SO_init}
\end{table*}
\begin{table*}
\begin{center}
\begin{tabular}{l|c|cccc|cccc|c}
\toprule
\multirow{3}{*}{Learning Method} & Hardness & \multicolumn{4}{c|}{Subject/Object} & \multicolumn{4}{c|}{Relation} & Combined\\
& Aware & many & medium & few & all & many & medium & few & all & all\\
& Diffusion & 86 & 255 & 1362 & 1703 & 16 & 46 & 248 & 310 & 2013\\
\midrule
\rowcolor{lightgray!50}\multicolumn{11}{l}{LSVRU \cite{zhang2019largescale}}\\
\multirow{2}{*}{CE + DiffAugment} & \ding{55} & \underline{63.18} & 41.96 & 9.58 & 17.14 & \underline{36.05} & 23.52 & \underline{6.13} & \textbf{10.26} & 13.70\\
& \ding{51} & 62.12 & \underline{42.42} & \underline{10.18} & \textbf{17.63} & 34.67 & \underline{23.73} & 6.04 & 10.14 & \textbf{13.88}\\
\hline
\rowcolor{lightgray!50}\multicolumn{11}{l}{RelTransformer \cite{chen2022reltransformer}}\\
\multirow{2}{*}{CE + DiffAugment} & \ding{55} & 70.14 & \underline{51.28} & \underline{14.27} & \textbf{22.64} & 51.56 & \underline{25.91} & 6.94 & 12.06 & 17.35\\
& \ding{51} & \underline{70.21} & 51.20 & 14.23 & 22.59 & \underline{53.15} & 25.56 & \underline{7.52} & \textbf{12.56} & \textbf{17.58}\\
\bottomrule
\end{tabular}
\end{center}
\caption{\textbf{Effect of using Hardness aware diffusion}. Combined accuracy refers to the average of all subjects/objects accuracy and all relations accuracy. For all the experiments, Random Gaussian seed has been used while sampling. The overall best results are in bold. Category-wise best results are underlined. For both the VRR approaches, Hardness aware diffusion gives better combined performance.}
\label{Table:Ablation_Hardness_diffusion}
\end{table*}
\begin{table*}
\begin{center}
\begin{adjustbox}{max width=\textwidth}
\begin{tabular}{l|c|cccc|cccc|c}
\toprule
\multirow{3}{*}{Learning Method} & Fine-tuning & \multicolumn{4}{c|}{Subject/Object} & \multicolumn{4}{c|}{Relation} & Combined\\
& strategy & many & medium & few & all & many & medium & few & all & all\\
&  & 86 & 255 & 1362 & 1703 & 16 & 46 & 248 & 310 & 2013\\
\midrule
\rowcolor{lightgray!50}\multicolumn{11}{l}{LSVRU \cite{zhang2019largescale}}\\
\multirow{2}{*}{WCE + DiffAugment} & Random & \underline{39.83} & \underline{33.90} & 17.33 & \textbf{20.95} & 37.37 & \underline{37.72} & 18.13 & 22.03 & 21.49\\
& Easy then Hard & 37.34 & 31.43 & \underline{17.74} & 20.78 & \underline{38.39} & 37.56 & \underline{18.64} & \textbf{22.47} & \textbf{21.62}\\
\hline
\rowcolor{lightgray!50}\multicolumn{11}{l}{RelTransformer \cite{chen2022reltransformer}}\\
\multirow{2}{*}{CE + DiffAugment} & Random & \underline{70.55} & 50.47 & 13.23 & 21.70 & 47.89 & \underline{30.13} & \underline{7.09} & \textbf{12.62} & 17.16\\
& Easy then Hard & 70.05 & \underline{51.10} & \underline{14.00} & \textbf{22.39} & \underline{55.38} & 24.65 & 6.97 & 12.09 & \textbf{17.24}\\
\bottomrule
\end{tabular}
\end{adjustbox}
\end{center}
\caption{\textbf{Effect of using Curriculum based fine-tuning} \ie initially using easy samples and later followed by hard samples. Combined accuracy refers to the average of all subjects/objects accuracy and all relations accuracy. For all the experiments, Subject-Object based seed has been used while sampling. The overall best results are in bold. Category-wise best results are underlined. For both the settings, curriculum based fine-tuning gives better combined performance.}
\label{Table:Ablation_Hardness_finetuning}
\end{table*}
\begin{table}
\begin{center}
\begin{tabular}{c|c|c|c}
\toprule
\multicolumn{4}{c}{LSVRU\cite{zhang2019largescale} + CE loss : 12.25\%}\\
\hline
Sbj/Obj & Hardness & Curriculum & Combined\\
based & Aware & based & accuracy\\
seed & Diffusion & Fine-tuning &\\
\midrule
\ding{55} & \ding{55} & \ding{55} & 13.7\\
\ding{55} & \ding{51} & \ding{55} & 13.88\\
\ding{55} & \ding{51} & \ding{51} & 13.99\\
\ding{51} & \ding{55} & \ding{55} & 13.62\\
\rowcolor{lightgray!50}\ding{51} & \ding{51} & \ding{55} & \textbf{14.66}\\
\ding{51} & \ding{51} & \ding{51} & 14.2\\
\bottomrule
\end{tabular}
\end{center}
\caption{Effect of the DiffAugment enhancements in different combinations on LSVRU \cite{zhang2019largescale} trained with Cross Entropy loss on the GQA-LT dataset. The best enhancement combination is highlighted in gray. Combined accuracy refers to the average of all subjects/objects accuracy and all relations accuracy.}
\vspace{-0.2cm}
\label{Table:Ablation_combined}
\end{table}
\begin{table}
\begin{center}
\begin{tabular}{c|c|c|c}
\toprule
\multicolumn{4}{c}{LSVRU\cite{zhang2019largescale} + CE loss : 12.25\%}\\
\hline
\#samples & Subject/Object & Relation & Combined\\
 & all & all & all\\
 & 1703 & 310 & 2013\\
\midrule
20K & 17.16 & 10.72 & 13.94\\
60K & 17.38 & 11.02 & 14.20\\
\rowcolor{lightgray!50}96K & 17.51 & 11.80 & \textbf{14.66}\\
\bottomrule
\end{tabular}
\end{center}
\caption{Effect of using different number of augmented samples to fine-tune LSVRU \cite{zhang2019largescale} trained with Cross-Entropy Loss on GQA-LT dataset. The best performance is highlighted in gray.}
\vspace{-0.2cm}
\label{Table:Ablation_num_aug_samples}
\end{table}
\subsection{Quantitative Results}
Table \ref{Table:Quantitative_Results} presents the results of our approach, including both the enhancements on the GQA-LT dataset. For LSVRU as well as RelTransformer, irrespective of the loss function used, fine-tuning using Diffusion generated samples improves the overall average per-class accuracy for both subjects/objects and relations. Further, Diffusion-augmented samples improve the performance of models trained using RelMix augmentation as well. As the table shows, DiffAugment based fine-tuning of an LSVRU model that has been trained with Cross Entropy and VilHub losses and using RelMix Augmentation, the overall combined performance increases by $1.42\%$. Even though the \textit{many} category performance drops, it is compensated by the considerable rise in the joint performance of the \textit{medium} and \textit{few} categories. The combined subject/object and relation performance increases by $\approx2.5\%$ for LSVRU \cite{zhang2019largescale} and $\approx1.1\%$ for RelTransformer \cite{chen2022reltransformer}.
\subsection{Ablation Study}
To validate the advantage of each of the enhancements, we perform extensive ablations and describe their takeaways as follows--\\ 
\noindent\textbf{Effect of using Subject-Object based seed} Table \ref{Table:Ablation_SO_init} shows the result of using subject-object based seed rather than random Gaussian seed while sampling from the Diffusion Model. Even though the overall subject/object performance reduces slightly, there is a consistent improvement in the overall relation performance for both the architectures (LSVRU \cite{zhang2019largescale} and RelTransformer \cite{chen2022reltransformer}) irrespective of the loss function used (CE/WCE). The improvement in the relations accuracy overcomes the slight reduction in subject-object accuracy as displayed by the better combined performance. For LSVRU, Subject-Object based seed improves the combined accuracy by $0.78\%$ and for RelTransformer, it increases the accuracy by $0.2\%$.\\
\noindent\textbf{Effect of Hardness aware diffusion} Table \ref{Table:Ablation_Hardness_diffusion} reports the change in accuracy across different categories on adding hardness as an extra condition to the Diffusion Model. Even though the category-wise performance for both relations and subjects/objects does not improve consistently, the combined average per-class accuracy across relations and objects improves for both LSVRU and Reltransformer.\\
\noindent\textbf{Effect of Curriculum based fine-tuning} Since we use hardness as a condition while generating the visual embeddings and inspired from the principles of curriculum learning \cite{soviany2022curriculum}, we try to observe if ordering the visual embeddings according to their hardness and using the easier samples first for fine-tuning the VRR model has any impact on the performance. So, we divide the triplets between easy and hard by calculating the entropy of the hardness vectors and choosing the median as the threshold. As observed in Table \ref{Table:Ablation_Hardness_finetuning}, there is no consistent rise/decline in category-wise accuracy for subjects/objects or relations even though the overall performance combined across all relations and objects improves. However, as seen in Table \ref{Table:Ablation_combined}, for LSVRU with CE loss, the optimal performance is achieved without curriculum based fine-tuning.\\
\begin{figure*}[h]
\begin{center}
\includegraphics[scale=0.3]{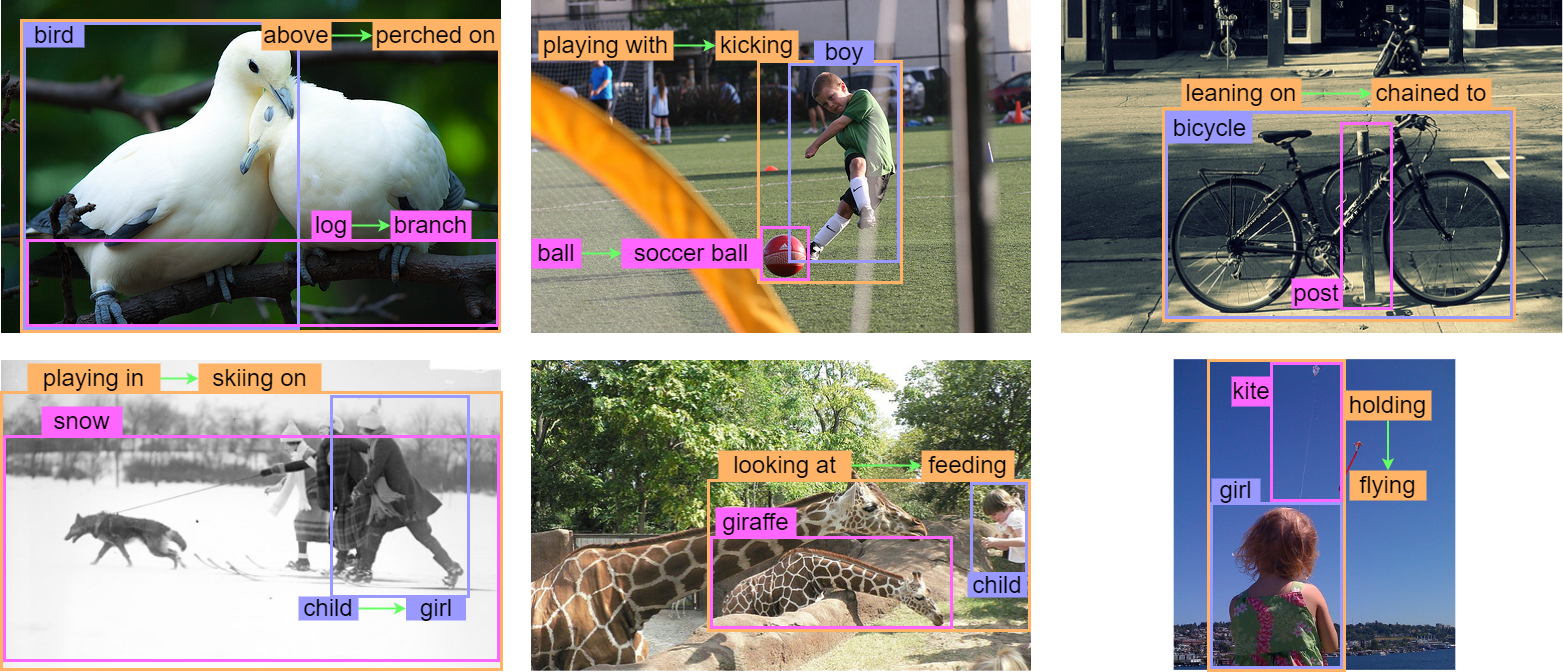}
\end{center}
   \caption{Visualizing Diffusion-Augmentation based results on GQA-LT dataset. Here, the green arrows show how the predicted subject/relation/object changes after fine-tuning LSVRU (with CE loss) model using DiffAugment. Orange color denotes the relation, blue denotes the subject, and pink denotes the object.}
\label{fig:qual_results}
\end{figure*}
\noindent\textbf{Combined Effect of all the enhancements} In order to understand the outcome of using all these enhancements together, we take the LSVRU \cite{zhang2019largescale} model trained using Cross-entropy loss as baseline and apply the Subject-Object seed based sampling, Hardness conditioned Diffusion and Curriculum based fine-tuning in all possible combinations. The overall relation + subject/object average per-class accuracy for each configuration is reported in Table \ref{Table:Ablation_combined}. As the table shows, adding each of the enhancements generally improves the overall accuracy, however, the optimal strategy may involve using only some of the enhancements (like not using curriculum based fine-tuning in this case).\\
\noindent\textbf{Effect of number of augmented samples used in fine-tuning} To observe how the number of DiffAugment generated visual embeddings affects the average per-class accuracy after fine-tuning, we create subsets of $20,000$ and $60,000$ samples from the entire set of $96,000$ embeddings (sampled using Subject-Object based seed and with Hardness conditional DM). Then we fine-tune the LSVRU model using Cross-Entropy loss on each of the subsets. The average per-class accuracy for relations, subjects/objects and both combined is displayed in Table \ref{Table:Ablation_num_aug_samples}. From the table, it is evident that more number of augmented samples result in better classification performance upon fine-tuning. However, the Diffusion sampling time that is required also increases for more samples.
\subsection{Qualitative Results}
We examine how the predicted S/R/O changes after fine-tuning an LSVRU + CE \cite{zhang2019largescale} model using DiffAugment and show six such results in Figure \ref{fig:qual_results}. It can be observed that the fine-tuned model is able to predict more informative relation labels such as \textit{perched on} instead of \textit{above}, \textit{kicking} rather than \textit{playing with}, \textit{chained to} instead of \textit{leaning on}, \textit{skiing on} rather than \textit{playing in}, \textit{feeding} instead of \textit{looking at} and \textit{flying} rather than \textit{holding}. All of these informative relations are more towards the tail of dataset distribution as compared to the initial predictions. DiffAugment can also make the subject/object more specific, as in \textit{log} changed to \textit{branch}, \textit{ball} changed to \textit{soccer ball} and \textit{child} changed to \textit{girl}.
\section{Conclusion}
We present DiffAugment - a pioneer in utilizing generative Diffusion Models to overcome the class imbalance of Long-tailed Visual Relationship Recognition. It is a streamlined approach to augment the tail classes by first generating viable tail class triplets with the help of WordNet \cite{10.1145/219717.219748}; and then generating visual embeddings corresponding to those triplets through Diffusion. We further define the \textit{hardness} of each triplet; utilize it as a diffusion conditional, and also introduce a new subject-object based seed for diffusion sampling -- all of which improve the discriminative performance of the VRR approaches upon fine-tuning as shown in the experiments. The offline design makes this method suitable to be used on top of any existing VRR model.
{
    \small
    \bibliographystyle{ieeenat_fullname}
    \bibliography{main}
}


\end{document}